\documentclass[12pt]{article} 
\usepackage{geometry}
\usepackage{graphicx}   
\usepackage{amsmath}    
\usepackage{hyperref}   
\usepackage[english]{babel}
\usepackage{amssymb}

\title{Stochastic Separability of Embedding Manifolds}
\author{Liqing Zhang\thanks{Email: zhang-lq@cs.sjtu.edu.cn}\\
       School of Computer Science \\
       Shanghai Jiao Tong University}
\date{\today}

\newtheorem{theorem}{Theorem}
\newtheorem{lemma}{Lemma}

\newcommand {\mbmu}{{\boldsymbol{\mu}}}
\newcommand {\mb}[1]{\boldsymbol{#1}}
\newcommand {\ww}{\boldsymbol{w}}
\newcommand{\Myproofc}{ \noindent \textbf {Proof}:\hskip 0.2cm }
\newcommand{\myendproof}{ \hfill \square}

\begin{document}

\maketitle

\begin{abstract}
Neurobiological studies and representation learning have observed that representations of objects belonging to the same category in high-dimensional neural spaces exhibit low-dimensional object manifold characteristics, and different  object manifolds are linearly separable in these neural spaces. However, these experimentally observed phenomena lack rigorous theoretical validation to date.

This paper proposes a new stochastic separability theorem for embedding manifolds of two different object categories. First, we establish a projection measure concentration theorem for embedding manifolds under general conditions. We develop a new two-layer measure concentration analysis technique, which unifies two  estimation bounds via the law of total expectation to derive measure concentration inequalities.

Based on the measure concentration theorem, we further prove a stochastic separability theorem for embedding manifolds of two different object categories. If two datasets have distinct means and bounded total variances, their samples become linearly separable with high probability, provided that the projection direction satisfies a non-singularity condition. 

The main contributions of this paper are twofold: 

1. We prove the projection concentration properties of embedding manifolds in high-dimensional spaces by using two-lawyer tail-bound inequalities.

2. We identify a non-singularity condition for the stochastic separability between embedding manifolds, and rigorously prove the stochastic projection separability theorem.

The theorem not only uncovers geometric and statistical properties of the object embedding manifolds, but also provides a novel mechanism for representation learning in deep networks. 
\end{abstract}



\section{ Introduction} 
In the era of deep learning and large foundation models, embedding multi-modal data into high-dimensional feature spaces has become a critical approach for machine learning modeling and inference. Although high-dimensional modeling confronts intricate challenges such as the curse of dimensionality, the interplay between randomness and distribution concentration in high-dimensional spaces can drastically simplify numerous high-dimensional machine learning problems—provided we relax error estimation requirements and only demand bounds to hold with high probability. This analytical paradigm greatly facilitates modeling and model evaluation for high-dimensional systems, such a positive effect of the high dimensionality is known as the blessing of dimensionality.

This idea was first proposed by Kainen \cite{BlessingDim:Kainen1997}, who noted that  having many parameters actually facilitates computation and uncovered intrinsic correlations between randomness and geometric properties of high-dimensional variables. Gorban \& Tyukin\cite{BelssingDim:Gorban2018mathematical} and Donoho \cite{BlessingDimDoniho2000} further refined and extended error bound estimation techniques for high-dimensional space models.

High-dimensional measure concentration theory not only reveals statistical distributions and geometric structures of high-dimensional data, but also furnishes new tools for high-dimensional modeling and generalization analysis. Its core modeling principle bears profound implications for understanding internal architectures and statistical regularities of deep learning models.

Measure concentration theory characterizes distribution properties of data following common distributions: uniform measures on unit balls/cubes, and Gaussian distributions. The concept of measure concentration originated with Milman \cite{Concentration1971milman:UnitSphere,Concentration1971milmannew}, who revisited Paul Lévy’s spherical isoperimetric inequalities within local Banach space theory and formalized spherical measure concentration as a universal mathematical phenomenon. Subsequent work by Ledoux et al.  \cite{ledoux1991probability,ledoux1996talagrand} generalized proof strategies for uniform spherical concentration, refining concentration bounds via logarithmic Sobolev inequalities \cite{gross1975logarithmic} and entropy methods \cite{ledoux1996talagrand}.

\begin{figure}
    \centering
    \includegraphics[width=0.40\linewidth]{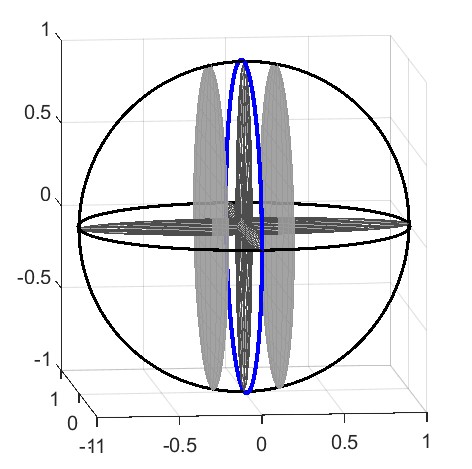}
    \caption{Illustration of equatorial/waist concentration: The mass of the unit ball concentrates in the thin layer of the blue cross-section}
    \label{fig:EquatorConcentration}
\end{figure}

Measure concentration on the unit sphere manifests in two key ways. First, nearly all volume mass of an n-dimensional unit ball $ \mathbb{B}^n $ is concentrated within a very thin shell adjacent to the sphere’s surface, while the central core of the unit ball occupies a volume that vanishes exponentially.
Second, mass of the unit ball accumulates in thin equatorial slices passing through the origin, as illustrated in Figure \ref{fig:EquatorConcentration}.  For any fixed $ \epsilon>0 $, the fraction of mass contained within a $ 2\epsilon $-thick slice converges to $1$ as the dimension tends to infinity. This property is known as equatorial/waist concentration, which provides the foundation for proving the embedding manifold concentration in this work.

Different random variables exhibit disparate concentration behavior. In $ \mathbb{R}^n $, the Euclidean norm $ \|X\|_2 $ of a standard Gaussian random vector concentrates tightly around $ \sqrt{n} $, forming the so-called Gaussian annulus \cite{vershynin2026high}:
\begin{equation}
p \left( \left| \frac{\|X\|_2}{\sqrt{n}} - 1 \right| \geqslant \epsilon \right) \leqslant 2 \exp\left(- c n \epsilon^2\right) , \quad \epsilon \in (0,1),
\end{equation}  
where constant $ c>0 $ and $ X\sim\mathcal{N}(0, \mb I_n) $.

While isotropic Gaussian vectors concentrate within a spherical annulus, Lipschitz functions defined on Gaussian vectors concentrate around their expectation $ \mathbb{E}[f(X)] $, with tail behavior fully controllable—analogous to univariate Gaussian random variables with well-behaved mean and variance bounds.

A complementary property tied to measure concentration is probabilistic convexity of random high-dimensional samples, formalized as the stochastic separation theorem by Gorban \& Tyukin \cite{gorban2017stochastic}  in 2017. This theorem reveals a counterintuitive high-probability linear separability in high dimensional spaces: under uniform distribution assumptions, any single sample within a finite dataset can be linearly separated from all remaining samples with arbitrarily high probability when the ambient dimension is sufficiently large.

Let $  X \in \mathbb R^{n} $ denote an n-dimensional random vector with independent, centrally symmetric components and bounded second moments. Let $\mathcal D =\big\{ \mb x_k \big\}_{k=1}^N$ denote $N$ independent identically distributed (i.i.d.) samples of $X$. For arbitrarily small $ \epsilon \in(0,1)  $, there exists a threshold dimension such that any single sample $ \mb x_k  $ can be Fisher-separated from the rest of the dataset with probability at least $ 1-\epsilon  $. Fisher separability requires the existence of a weight vector $\mb w$  satisfying $\langle \mb w,\mb x_k\rangle>\max_{i\ne k}\langle \mb w,\mb x_i\rangle$ , equivalent to existence of a separating linear hyperplane for $  \mb x_k $.

Grechuk et al. (2021)  \cite{grechuk2021general} extended this result to broader distribution families and derived tight finite-dimensional separation probability bounds. Their work relaxes symmetry and sub-Gaussian constraints from the original theorem, generalizing stochastic separation to log-concave, mixture, and clustered non-independent datasets with computable upper/lower bounds on separation probability. The generalized stochastic separation theorem is applied to correct errors and analyze instabilities in artificial intelligence systems.

Stochastic separation theorems provide a number of potential applications in the field of machine learning, such as  sample correction, intrinsic dimension estimation, and few-shot learning. Gorban et al. \cite{gorban2018correction} \cite{gorban2021highdimensional} leveraged Fisher discriminants for one-shot learning of model correctors, with stochastic separability furnishing complete mathematical foundations for such frameworks. For fine-grained clustered data, A novel stochastic separation theorem is developed  for designing a multi-corrector framework validated on deep convolutional neural network error correction and novel class incremental learning using the CIFAR-10 benchmark.

Albergante et al. \cite{SSThm-albergante2019estimating} proposed an intrinsic dimension estimation algorithm using Fisher separability from stochastic separation theory, matching state-of-the-art performance across standard and real biological datasets with efficient computation and robustness to noisy samples. Tyukin et al. \cite{SSThm-tyukin2019highdimensionalbrain} investigated information processing properties of convergent neural pathways onto single neurons, indicating that via the measure concentration and the stochastic separation models, individual neurons can selectively encode, recognize, and memorize arbitrary information units through associative dynamic learning of new content.

Classical measure concentration theory typically assumes uniform supported on bounded domains or Gaussian distributions. Although stochastic separation theorems characterize probabilistic convexity and separability within a single dataset, they cannot be applied directly to separability between two distinct real-world datasets. To characterize separability between sample sets of two distinct object classes in the embedded feature spaces, we first analyze distribution properties of samples belonging to a single object category.

Samples of images from same-class objects do not follow uniform distributions in high-dimensional embedding spaces. For computer vision, images of the same object share structural attributes; continuous pose variation induces continuous shifts in embedding representations, forming low-dimensional sub-manifolds within feature spaces. For example, neural population responses to images of cats versus dogs, each form distinct object manifolds, reformulating object recognition to a separation problem between pairs of embedding manifolds.

Since latent parameters governing same-category images are finite-dimensional, the intrinsic dimension of the embedding manifold of same-category object images remains bounded and does not grow to infinity with ambient feature space dimension. In addition, Noise and background interference prevent these distributions from forming perfect mathematical manifolds—instead, samples cluster within neighborhoods of the underlying manifold, which we term stochastic embedding manifolds.

Low-dimensional manifold embedding representations for same-category object images are not merely theoretical machine learning assumptions; they are substantiated by extensive neurobiological and systems biology experiments. Research across neuroscience, single-cell transcriptomics, and metabolomics confirms pervasive structural redundancy in high-dimensional biological observational data, where functional states and temporal evolution are governed by a small set of latent variables, naturally inducing low-dimensional manifold distributions.

Within visual hierarchical systems, neuronal population responses evoked by identical objects under varying pose collectively form object manifolds. DiCarlo \& Cox \cite{dicarlo2007untangling}  integrated neurophysiology and computational theory to define object manifolds as continuous surfaces traced by neuronal responses to objects under translation, scaling, and pose variation in high-dimensional spaces. Each transformation layer within the ventral visual stream aims to disentangle object manifolds to enable rapid visual categorization, converting initially non-separable manifolds into linearly separable structures.

This embedding manifold representation is corroborated by studies of primary visual cortex and motor cortex neural representations \cite{dicarlo2012brain, pagan2013signals, grootswagers2019untangling, henaff2019perceptual} , as well as deep neural network representation learning literature\cite{ranzato2007unsupervised, goodfellow2009measuring,bengio2013representation}. The perceptual manifold geometric frameworks \cite{chung2018classification,cohen2020separability} quantify three measurable manifold attributes: intrinsic manifold dimension, manifold radius, and inter-manifold correlation. Separability between distinct perceptual manifolds depends jointly on intrinsic dimension, radius, and cross-manifold correlation—this theory provides  analytical tools to explain manifold disentanglement and improved separability within deep network representations. Chung \& Abbott \cite{CHUNG2021NeuralGeo} demonstrated perceptual manifold properties extend beyond visual stimulus encoding to motor and cognitive brain regions. They systematically investigated neural manifold methods across perceptual disentanglement, manifold classification theory, abstract encoding in cognitive systems, topological cognitive maps, and dynamic motor disentanglement, formalizing mathematical links between geometric representation properties and encoded task information.

In summary, neurobiological experiments have revealed that same-category objects induce low-dimensional manifold neural representations in high-dimensional deep brain spaces, with distinct object manifolds linearly separable at deeper layers. Deep embedding experiments similarly show that well-trained representation models produce embedding manifolds for separate object classes that become linearly separable with high probability, with the separation probability rising alongside the ambient feature dimensionality.

Despite consistent experimental evidence for low-dimensional embedding manifolds and high-probability separability between cross-class manifolds, rigorous theoretical proofs remain absent in the prior literature. This paper derives sufficient projection concentration conditions for the embedding manifold of same-category objects and proves the projection measure concentration under general conditions. For two-class separability between stochastic embedding manifolds, we prove that if two datasets have distinct means and bounded total variances, their embedding manifolds become linearly separable with arbitrarily high probability when ambient feature dimension is sufficiently large—provided projection directions satisfy a non-singularity criterion. We formalize this result as the embedding manifold stochastic separability theorem.

Although the theorem states that linear separability improves with embedding dimension, higher dimensionality does not universally imply  superior model performance, which must be evaluated holistically via generalization and interpretability metrics. The stochastic separability theorem delivers a new mechanism for representation learning: minimize the intrinsic dimension of each object’s manifold during embedding learning. Additionally, high-dimensional embeddings allow the deployment of simple linear classifiers for downstream categorization, reducing overfitting risks.

\section{Projection Measure Concentration of Embedding Manifolds}
High-dimensional embeddings are fundamental to deep learning and generative artificial intelligence. The curse of dimensionality renders most low-dimensional data algorithms computationally intractable in high dimensions, yet the unique geometric and statistical properties of high-dimensional spaces unlock novel analytical frameworks for statistical learning and inference. Randomness induced by high ambient dimensionality enables innovative methodologies for high-dimensional data analysis.

Let $X=(x_1,\cdots,x_n)^\top \in \mathbb R^{n} $ denote an n-dimensional random vector with independent components, mean $\mbmu_X$, and component variances $\left\{ \sigma^2_i \right\}_{i=1}^n$.  As established in Section 1, samples of natural objects lie within neighborhoods of low-dimensional embedding manifolds whose intrinsic dimension remains finite and independent of ambient feature dimension. We characterize each stochastic embedding manifold via a probability density function $p_X(\mb x)$ describing the distribution of samples drawn from the manifold. Bounded intrinsic dimensionality of a stochastic embedding manifold is captured via a total variance constraint on the n-dimensional random vector X: there exists a constant $ C_0 $ such that
\begin{equation}  
\sum_{i=1}^n \sigma_i^2 \leqslant C_0, \quad \forall \ n .
\end{equation}  
We term this the bounded total variance condition. Let $\mathcal D_X=\big\{ \mb x_k \big\}_{k=1}^N $ denotes $N$ i.i.d. samples of $X$ distributed over the stochastic embedding manifold. We use $ \mathbb{N}_+ $ for positive integers and $\mathbb S^{n-1}$to denote the unit n-dimensional sphere.

\begin{theorem} [Projection Measure Concentration for Embedding Manifolds]  \label{thmManifoldConcen}
Let X be an n-dimensional random vector with independent components, mean $ \mbmu_X $, and bounded total variance. For arbitrary $ t>0 $, $ \epsilon,\delta\in(0,1) $, and any projection vector $\ww \in \mathbb S^{n-1}$, there exists $ n_0\in\mathbb{N}_+ $ such that for all $ n>n_0 $, the inequality below holds with probability at least $ 1-\delta $:
\begin{equation}  \label{eq:ThmConcen}
p\left( \left\{ \big| \ww^\top( X-\mbmu_X) \big| \geqslant t \right\} \right)  \leqslant \epsilon  .
\end{equation}  
\end{theorem}

\Myproofc
Given a projection vector $ \ww\in \mathbb S^{n-1}$, consider the random variable $ \ww^\top(X-\mbmu_X) $. By Chebyshev’s inequality, for any $ t>0 $:
\begin{equation}  \label{Ch3:RandomTwosetX}
p\left( \left| \ww^\top (X-\mbmu_X)\right|  \geqslant t  \right) \leqslant \frac{\mathbb{V}ar  [\ww^\top (X-\mbmu_X) ]}{t^2},
\end{equation} 
where the variance term $\mathbb{V}ar  \left[\ww^\top (X-\mbmu_X) \right]=\sum_{i=1}^n w_i^2 \sigma^2_i$ and $ \ww^\top \ww=1 $. We next apply measure concentration properties of uniformly random vectors $\ww$  on $\mathbb  S^{n-1}$ to bound the random quantity $Z_w=\mathbb{V}ar  \left[\ww^\top (X-\mbmu_X)\right]$.  Derivations in the appendix yield moment identities for spherical components:
\begin{equation}  
\mathbb{E} [w_i^2] = \frac 1n; \qquad  \mathbb{E} [w_i^4]=\frac{3}{n(n+2)}.
\end{equation}  
The expectation and variance of $Z_w=\mathbb{V}ar  \left[\ww^\top (X-\mbmu_X)\right]$ are computed as:
\begin{eqnarray}  \label{MeanZX}
\mathbb{E} [Z_w] &=& \frac 1n \sum_{i=1}^n \sigma^2_i \overset{\triangle}{=\! \! =} \frac 1n C_2 \\ 
\mathbb{V}ar  [Z_w] &=& \frac{2}{n(n+2)} \left(   \sum_{i=1}^n   \sigma_i^4 - \frac 1 n \Big(  \sum_{i=1}^n  \sigma_i^2 \Big) ^2\right) \leqslant  \frac{2}{n(n+2)} C_4 .
\end{eqnarray}   
with shorthand $C_2= \sum_{i=1}^n  \sigma^2_i , \  C_4=  \sum_{i=1}^n  \sigma^4_i $. Under the bounded total-variance condition, both $ C_2,C_4 $   are bounded constants, so $ \mathbb{E}[Z_w],\mathbb{V}ar   [Z_w]\to 0 $   as the dimension $n$ increases, implying $ Z_w $ converges to zero with high probability.

To estimate the tail bound of $Z_w$, we apply Chebyshev’s inequality to $ Z_w $, for any $ \eta>0 $, 
\begin{equation}   \label{eq:ZWBound}
p\big(|Z_w-\mathbb{E} [Z_w]|\geqslant \eta \big) \leqslant \frac{\mathbb{V}ar  [Z_w]}{\eta^2} \leqslant \frac{2}{n(n+2)} \frac{C_4}{\eta^2}.
\end{equation}  
Set $ \eta=1/\sqrt{n} $ to guarantee a tight concentration around $ \mathbb{E}[Z_w] $. There exists a threshold integer $ n_1 $ such that for all $ n\geqslant n_1 $,
\begin{equation}   \label{eq:ZWBound2}
\frac{2}{n(n+2)} \frac{C_4}{\eta^2} \leqslant \delta, \ \ \forall \ n \geqslant n_1. 
\end{equation}  
Substituting \eqref{eq:ZWBound2} into equation \eqref{eq:ZWBound} yields
\begin{equation} 
p\big(|Z_w-\mathbb{E} [Z_w]| \geqslant  1/\sqrt{n} \big) \leqslant \delta, \ \ \forall \ n \geqslant n_1.
\end{equation}  
Equivalently:
\begin{equation} 
p\big( | Z_w  -\mathbb{E} [Z_w]| <  1/\sqrt{n} \big) > 1- \delta , \ \ \forall \ n \geqslant n_1. 
\end{equation}  
Taking the one-side bound for random variable $Z_W$, we obtain 
\begin{equation}   \label{ZXMeanZX}
p\big( Z_w <  \mathbb{E} [Z_w]  +  1/\sqrt{n} \big) > 1- \delta.
\end{equation}  
Using the expectation identity \eqref{MeanZX}, we derive that there exists $ n_2\in\mathbb{N}_+ $ such that for all $ n\geqslant n_2 $:
\begin{equation}  \label{MZXBound}
\mathbb{E} [Z_w]  +  1/\sqrt{n} = \frac{C_2}{n}+\frac{1}{\sqrt{n}} \leqslant  t^2 \ \epsilon.
\end{equation}  
Define $ n_0=\max\{n_1,n_2\} $. Then combining two inequalities \eqref{ZXMeanZX} and \eqref{MZXBound} yields that for all $ n\geqslant n_0 $ 
\begin{equation}   \label{ZWlesstteps}
p\big( Z_w < {t^2 \ \epsilon}   \big) >  1- \delta.
\end{equation}  
We now establish event inclusion relations to combine the dual probability bounds. Define three events:
\begin{eqnarray}  
E_0 &=&\big \{ \ww \mid p\left( \left| \ww^\top (X-\mbmu_X)\right| \geqslant t  \right) \leqslant \epsilon  \big \}, \nonumber  \\
E_1 &=& \big \{\ww \mid  p\left( \left| \ww^\top (X-\mbmu_X)\right| \geqslant t  \right) \leqslant Z_w \big / t^2  \big \}, \nonumber \\
E_2 &=& \big \{ \ww \mid  Z_w < {t^2 \ \epsilon}  \big \} . \nonumber
\end{eqnarray}  
The conjunction $ E_1\cap E_2 $ implies $ E_0 $, i.e., $ E_0\supseteq E_1\cap E_2 $. Under the theorem conditions, $ E_1 $ holds for all $ \ww\in S^{n-1} $ , thus we obtain 
\begin{equation}   \label{EXConfidence}
p(E_0) \geqslant p(E_1 \cap E_2 ) =  p(E_2)  > 1-\delta .
\end{equation} 
This confirms the following target bound holds with probability at least $ 1-\delta $  
\begin{equation} 
 p\left( \left| \ww^\top (X-\mbmu_X)\right| \geqslant t  \right) \leqslant \epsilon.
\end{equation} 
Thus, this completes the proof.   $\myendproof$

The proof employs two nested layers of measure concentration estimation: first, Chebyshev's inequality for $ \ww^\top(X-\mbmu_X) $ provides a tail bound of projected data. The tail bound $ Z_w $ depends on spherical projection $\ww$. Second, we apply concentration inequalities to $ Z_w $ itself, showing its mean and variance vanish as the dimension $n$ goes to infinity, resulting in $ Z_w < t^2\epsilon $ with high probability for sufficiently large $n$.

We name the set $ E_0 $ the set of good projection samples $\ww$, since arbitrary small $ \delta $ ensures random vector $\ww$ falls within $ E_0 $ with probability at least $ 1-\delta $. All good sample projections satisfy the projection concentration bound \eqref{eq:ThmConcen}. The complement $ E_0^c $ contains bad sample projections that violate the concentration inequality. The typical bad vectors are the standard coordinate vectors $ \mb e_i $ for each coordinate axis—projection onto a single coordinate axis recovers the marginal distribution of $ X_i $, which lacks the measure concentration property.

We can reformulate the tail bound \eqref{eq:ThmConcen} of Theorem \ref{thmManifoldConcen} into its equivalent form: for any random sampled $\ww$, fixed $ t>0 $, and arbitrarily small $ \epsilon,\delta\in(0,1) $:
\begin{equation} 
 p\left( \left| \ww^\top (X-\mbmu_X)\right| <  t  \right) >  1-\epsilon .
\end{equation} 
This highlights a critical distinction between projection concentration of bounded-total-variance embedding manifolds and isotropic Gaussian distributions: projections of isotropic Gaussian vectors remain univariate Gaussian and do not concentrate within arbitrarily small intervals around their projection mean. Under the condition of the bounded total-variance, embedding manifold projections concentrate within arbitrarily narrow bands around the projected mean for sufficiently high ambient dimension $n$, for nearly all random projection directions $\ww$.

Theorem \ref{thmManifoldConcen} describes the tail bound for measure concentration  in the nested dual-probability way.  We next derive a lemma converting dual probability bounds into single-layer concentration inequalities.

\begin{lemma} \label{lemmaNested}
Suppose conditional tail bound $ p(|f(X,W)| \geqslant t \mid W=w) \leqslant g(w) \ ( \leqslant 1 )$, and $ p(g(W) \le \epsilon) \ge 1 - \delta $, then
\begin{equation}  
p(|f(X,W)| \geqslant t) \leqslant \epsilon + \delta. 
\end{equation}  
\end{lemma}

\Myproofc
Apply the law of total expectation to the tail event $E=\big\{ |f(X,W)| \geqslant t \big\} $:
\begin{equation}   
p(|f(X,W)| \geqslant t) = \mathbb{E} _{W}\big[ p(|f(X,W)| \geqslant t \big| \ W ) \big].
\end{equation}  
By the conditional bound $ p(|f(X,W)| \geqslant t \mid W=w) \leqslant g(w)$, we obtain 
\begin{equation}   
p(|f(X,W)| \geqslant t) = \mathbb{E} _{W}\big[ p(|f(X,W)| \geqslant t \mid W ) \big] \leqslant \mathbb{E} _{W}\big[ g(W) ) \big]  .
\end{equation}  
Decompose the expectation over disjoint indicator events $ \{g(W)\leqslant \epsilon\} $ and $ \{g(W)>\epsilon\} $:
\begin{eqnarray}  
p\big(|f(X,W)| \geqslant t \big)   \leqslant  
\mathbb{E} \big[g(W) \cdot \mathbf{1}_{\{g(W) \leqslant \epsilon\}} \big]  
 + 
\mathbb{E}\big[ g(W) \cdot \mathbf{1}_{\{g(W) > \epsilon\}} \big].
\end{eqnarray}   
The first term is bounded via $ g(W)\leqslant \epsilon $ on its support $\mathbf{1}_{\{g(W) \leqslant \epsilon\}}$:
\begin{equation}  
\mathbb{E}\big[ g(W)  \cdot \mathbf{1}_{\{g(W) \le \epsilon\}} \big]
\le \epsilon p(g(W)\leqslant \epsilon ).
\end{equation}  
The second term uses the trivial upper bound $ g(W)\leqslant 1 $:
\begin{equation} 
\mathbb{E}\big[g(W) \cdot \mathbf{1}_{\{g(W) > \epsilon\}} \big]
\le 1 \cdot p\big(g(W) > \epsilon \big).
\end{equation}  
Combining the  above two estimates, 
\begin{equation}  
p\big( |f(X,W)| \geqslant t \big)  \leqslant \epsilon \cdot p(g(W) \le \epsilon) + p(g(W) > \epsilon).
\end{equation}  
The lemma's premise $ p(g(W) \le \epsilon) \ge 1 - \delta $ implies $ p(g(W) > \epsilon) \le \delta $, hence:
\begin{equation} 
p(|f(X,W)| \geqslant t) \le \epsilon \cdot 1 + \delta = \epsilon + \delta.
\end{equation}  
This completes the proof. $\myendproof$

\begin{theorem}  [ Projection Concentration Theorem]  \label{ThmStochasticProject}
Let $X$ be an n-dimensional random vector with independent components, mean $ \mbmu_X $, and bounded total variance. Given $ t>0,\epsilon\in(0,1) $, there exists   $ n_0 \in \mathbb N_+ $  such that for all $ n>n_0 $:
\begin{equation}   \label{eq:thmConcen}
p\left( \left\{ \big| W^\top( X-\mbmu_X) \big| \geqslant t \right\} \right)  \leqslant \epsilon  
\end{equation}  
\end{theorem}

\Myproofc
Define $f(X,W)=W^\top (X-\mbmu_X)$, $g(W)=\mathbb{V}ar  [W^\top (X-\mbmu_X)/t^2 $.  From Theorem \ref{thmManifoldConcen}'s proof, for arbitrary $ \epsilon_1,\delta_1\in(0,1) $, there exists $ n_0\in\mathbb{N}_+ $   satisfying:
\begin{eqnarray} 
& & p\left(  \big| f(X,W) \big| \geqslant t \mid W=\ww  \right)  \leqslant g(\ww) , \\
& & p\big( g(W) \leqslant \epsilon_1 \big) \geqslant 1- \delta_1 .
\end{eqnarray}   
If we choose $ \epsilon_1=\epsilon/2,\delta_1=\epsilon/2$ ,  directly applying  Lemma \ref{lemmaNested} yields the tail bound \eqref{eq:thmConcen}.  $\myendproof$ 

The tail bound inequality \eqref{eq:thmConcen} in Theorem \ref{ThmStochasticProject} can be rewritten in its equivalent  form:
\begin{equation} 
p\left( \left\{ \big| W^\top( X-\mbmu_X) \big| < t \right\} \right)  > 1- \epsilon.
\end{equation} 
This inequality states that for arbitrarily small concentration radius t and error tolerance $ \epsilon $, projected samples concentrate within an arbitrarily narrow band around the projected mean with probability at least $ 1-\epsilon $, provided ambient embedding dimension $n$ is sufficiently large. The critical assumption is bounded total variance: indicating the low intrinsic dimensionality of embedding manifolds in high-dimensional spaces, should be independent of ambient dimension growth. From a high-dimensional perspective, low-dimensional manifolds occupy compact subspaces whose samples concentrate tightly around their global mean—a behavior observable via random linear projections. 

\section{Stochastic Projection Separability of Embedding Manifolds}  
The preceding section established projective measure concentration for single embedding manifolds under bounded total variance. We now prove that two distinct stochastic embedding manifolds with distinct means and bounded total variances  become linearly separable with high probability when ambient embedding dimension is sufficiently large.

Let $ X=(x_1,\dots,x_n)^\top\in\mathbb{R}^n $ denote an n-dimensional random vector with independent components, mean $ \mbmu_X $, component-wise variances $ \{\sigma_i^2\}_{i=1}^n $. Let $ Y=(y_1,\dots,y_n)^\top\in\mathbb{R}^n $   denote a second independent n-dimensional random vector with mean $ \mbmu_Y $, component-wise variances $ \{\rho_i^2\}_{i=1}^n $. Let $ \mathcal D_X=\{x_k\}_{k=1}^N $ denote N i.i.d. samples of X from its stochastic embedding manifold; let $ \mathcal D_Y=\{y_k\}_{k=1}^M $ denote M i.i.d. samples of Y from its manifold.

\begin{theorem}  [Stochastic Projection Concentration for embedding Manifolds]  \label{ThmSeparation} 
Suppose random vectors X,Y are component-independent, and satisfy bounded total variance and different means $ \mbmu_X\neq\mbmu_Y $. Given any  $ t>0,\ \epsilon,\ \delta\in (0,1) $ and any projection vector $ \mb w \in \mathbb S^{n-1} $, there exists $ n_0\in\mathbb{N}_+ $ such that for all $ n>n_0 $, the joint concentration event below holds with probability at least $ 1-\delta $:
\begin{equation} 
p\left( \left\{ \big| \ww^\top( X-\mbmu_X) \big| < t \right\} \cap \left\{  \big|\ww^\top (Y-\mbmu_Y)  \big| <t \right\} , \ \forall X\in \mathcal D_X,\ Y\in \mathcal D_Y \right) > 1-\epsilon . 
\end{equation}  
\end{theorem}
\Myproofc 
Apply Theorem \ref{thmManifoldConcen} separately to $X$ and $Y$. Given $ t>0, \ \epsilon_1=\epsilon/2, \ \delta_1 = \delta/2 \in(0,1) $, there exist dimension thresholds $ n_1,n_2\in\mathbb{N}_+ $ such that the following inequalities hold with probability at least $1-\delta_1$
\begin{eqnarray} 
& &  p \Big(\big|\ww^\top(X-\mbmu_X)\big|\geqslant t\Big)\leqslant \epsilon_1, \quad \forall n \geqslant n_1,  \\ 
& &  p \Big(\big|\ww^\top(Y-\mbmu_Y)\big|\geqslant t\Big)\leqslant \epsilon_1, \quad \forall n \geqslant n_2. 
\end{eqnarray}  
Define good-projection events for each manifold, 
\[
\begin{aligned}
E_X &= \Big\{ \ww \,\Big|\,  p \big(\big|\ww^\top(X-\mbmu_X)\big|\geqslant t\big)\leqslant \epsilon_1\Big\},\\
E_Y &= \Big\{ \ww \,\Big|\,  p \big(\big|\ww^\top(Y-\mbmu_Y)\big|\geqslant t\big)\leqslant \epsilon_1\Big\}.
\end{aligned}
\]
Using the probability of union bounds, we have 
\begin{equation} 
 p (E_X\cap E_Y)\geqslant  p (E_X)+ p (E_Y)-1 > (1-\delta_1)+(1-\delta_1)-1 = 1-\delta. \end{equation} 
When both $ E_X, \ E_Y $ hold simultaneously, apply complement probability identities for joint events. Set $ n_0=\max\{n_1,n_2\} $. For all $ n\geqslant n_0 $, 
\begin{eqnarray}  
& & p\Big(\big\{\big|\ww^\top(X-\mbmu_X)\big|<t\big\}\cap\big\{\big|\ww^\top(Y-\mbmu_Y)\big|<t\big\}\Big) \nonumber \\
& & \quad =1-p\Big(\neg\big\{\big|\ww^\top(X-\mbmu_X)\big|<t\big\}\cup\neg\big\{\big|\ww^\top(Y-\mbmu_Y)\big|<t\big\}\Big) \nonumber \\
& & \quad \geqslant 1-\Big[ p \big(\big|\ww^\top(X-\mbmu_X)\big|\geqslant t\big)+ p \big(\big|\ww^\top(Y-\mbmu_Y)\big|\geqslant t\big)\Big] \nonumber \\
& & \quad >1-(\epsilon_1+\epsilon_1)=1-\epsilon.
\end{eqnarray}   
This completes the proof.  $\myendproof$

This theorem infers a general result under mild conditions. If two datasets satisfy the bounded total variance, projections of two embedding manifolds concentrate tightly around their corresponding projected means with high probability provided the ambient dimension n is sufficiently large. Although low-dimensional embedding manifolds can be projected into narrow equatorial disks centered at their projected means, the marginal projection bands may overlap. 
Joint concentration alone does not guarantee linear separability between the two classes, an additional non-singularity condition is required:
\begin{equation} 
    \ww^\top(\mbmu_X-\mbmu_Y)\neq 0. 
\end{equation}  
This condition enforces separation between the two projected cluster centroids, termed the non-singular projection condition.

\begin{theorem} [Stochastic Projection Separability Theorem]  \label{thm:Separable} 
Let X,Y denote component-independent random vectors with bounded total variance and distinct means $\mbmu_X \neq\mbmu_Y $. Given $ \epsilon, \ \delta \in (0,1) $, and any non-singular projection vector $  \ww\in \mathbb S^{n-1} $   satisfying $ \ww^\top(\mbmu_X-\mbmu_Y)\neq0 $, there exists $ n_0 \in\mathbb{N}_+ $ such that for all $ n>n_0 $, the following inequalities hold with probability at least $1-\delta$, 
\begin{eqnarray} 
p\left( \left\{\ww^\top X > \ww^\top Y \right\} \right) \geqslant 1-\epsilon, \quad \text{if} \ \ \ww ^\top (\mbmu_X -\mbmu_Y) > 0 ;  \\ 
p\left( \left\{\ww^\top X < \ww^\top Y \right\} \right) \geqslant 1-\epsilon, \quad \text{if} \ \ \ww ^\top (\mbmu_X -\mbmu_Y) <  0 . 
\end{eqnarray}   
\end{theorem}

\Myproofc \hskip 0.1cm 
Without loss of generality, assume $ \ww^\top(\mbmu_X-\mbmu_Y)>0 $. Define the concentration radius $ t=\ww^\top(\mbmu_X-\mbmu_Y)/2 $. By Theorem \ref{ThmSeparation} there exists threshold dimension $ n_0 $ such that for all $ n>n_0 $,  the following inequality holds with probability at least $1-\delta$, 
\begin{equation}  \label{eq:twoConcen}
p\Big(\big\{\big|\ww^\top(X-\mbmu_X)\big|<t\big\}\cap\big\{\big|\ww^\top(Y-\mbmu_Y)\big|<t\big\},\;\forall X\in\mathcal{D}_X,Y\in\mathcal{D}_Y\Big)>1-\epsilon. 
\end{equation} 
The joint concentration event implies two strict inequalities:
\begin{eqnarray}  
&& \ww^\top X > \ww^\top\mbmu_X - t = \ww^\top\cdot\frac{\mbmu_X+\mbmu_Y}{2}, 
\\
&& \ww^\top Y < \ww^\top\mbmu_Y + t = \ww^\top\cdot\frac{\mbmu_X+\mbmu_Y}{2}. 
\end{eqnarray}   
Combining these inequalities yields $ \ww^\top X > \ww^\top Y $. We infer the following event inclusion relation,
\begin{equation}  
\big\{\ww^\top X>\ww^\top Y\big\}\supseteq \big\{\big|\ww^\top(X-\mbmu_X)\big|<t\big\}\cap\big\{\big|\ww^\top(Y-\mbmu_Y)\big|<t\big\}.
\end{equation}  
Thus, the corresponding probabilities satisfy the following inequality 
\begin{equation} 
p\big(\ww^\top X>\ww^\top Y\big)\geqslant p \Big(\big\{\big|\ww^\top(X-\mbmu_X)\big|<t\big\}\cap\big\{\big|\ww^\top(Y-\mbmu_Y)\big|<t\big\}\Big)>1-\epsilon. 
\end{equation} 
On the other hand, we define two event families to quantify confidence bounds:
\begin{eqnarray}  
E_C &\triangleq & \Big\{\ww \,\Big|\,  p \Big(\big\{\big|\ww^\top(X-\mbmu_X)\big|<t\big\}\cap\big\{\big|\ww^\top(Y-\mbmu_Y)\big|<t\big\}\Big)>1-\epsilon\Big\}, \\
E_S  &\triangleq & \big\{\ww \,\big|\,  p (\ww^\top X>\ww^\top Y)>1-\epsilon\big\}.
\end{eqnarray}   
Any projection vector $ \ww\in E_C $ also satisfies the separability bound $p (\ww^\top X>\ww^\top Y)>1-\epsilon$ in $E_S$, that is $ E_C\subseteq E_S $, implying $  p (E_S)\geqslant p (E_C)>1-\delta $. Thus the conclusion follows.  $\myendproof$

This theorem proposes general sufficient conditions for linear separability between two embedding manifolds: bounded total variance for each class and non-singular projection direction. Under the conditions of theorem \ref{thm:Separable}, given a target classification accuracy $ 1-\epsilon $, the theorem indicates that the two-class data are linearly separable with probability at least $ 1-\epsilon $ along good non-singular projection $\ww$, provided the ambient embedding
dimension $n$ is sufficiently large. From the proof of theorem \ref{ThmSeparation}, the probability of good projection samples is very high. We can easily find those good projection samples by combining the canonical projection vector with the random samples. We define a linear discriminant function as 
\begin{equation}  
\ell(X)=\ww^\top\Big(X-\frac{\mbmu_X+\mbmu_Y}{2}\Big).
\end{equation}  
The samples are assigned to class $ \mathcal D_X $ if $ \ell(X)>0 $, and to $ \mathcal D_Y $ if $ \ell(X)<0 $.

Two assumptions are irreplaceable: bounded total variance and projective non-singularity.
\begin{enumerate}
    \item 
    Bounded total variance enforces projective measure concentration for embedding manifolds. If this condition fails, projections do not concentrate within arbitrarily small intervals around the projected mean even as the dimension $n$ grows. For example, isotropic Gaussian distributions assign equal variance to every coordinate, violating bounded total variance. Accordingly, the projected data of isotropic Gaussian distributions follow a single-variable Gaussian distribution without tight concentration around the mean.
    \item 
    Projective non-singularity ensures the projected class centroids do not collapse onto identical points. Uniform random sampling of $\ww$ on $\mathbb S^{n-1} $ yields near-orthogonality with $ \mbmu_X-\mbmu_Y $ in high dimensions, i.e., $ \ww^\top(\mbmu_X-\mbmu_Y)\approx0 $, violating non-singularity. To construct valid random non-singular projections, first build a canonical separating direction, 
\begin{equation} 
\ww_0=\frac{\mbmu_X-\mbmu_Y}{\|\mbmu_X-\mbmu_Y\|_2},
\end{equation}  
which trivially satisfies $ \ww_0^\top(\mbmu_X-\mbmu_Y)>0 $. Then we can randomly generate $\mb \epsilon \sim \mathcal N(\mb 0, \mb \sigma^2_{\epsilon}) $, where $\mb \sigma^2_{\epsilon}$ is supposed to be bounded total-variance. Thus we construct valid projection vectors as $ \ww=\pm \ww_0+\mb \epsilon $, followed by unit normalization. Such constructed projection vectors are able to balance randomness and non-singularity.
\end{enumerate}

\section{ Discussions and Perspectives} 
The embedding vectors of images of same-category objects in high-dimensional feature spaces are located in the vicinity of a low-dimensional manifold, so the probabilistic distribution of the embedding vectors is not isotropic with the coordinate-wise variances. The stochastic separation theorem proposed by Gorban \& Tyukin \cite{gorban2017stochastic} deals with uniform distributed samples on unit spheres, whereas this paper addresses a more realistic two-distribution separability problem for object embedding manifolds.

This work addresses two core problems for stochastic embedding manifolds: projection measure concentration and cross-manifold stochastic separability. Different from the isotropic Gaussian distributions, where the projection of Gaussian random vector  becomes uni-variate Gaussian without tight concentration around the projected mean,  the probability distribution of stochastic embedding manifolds with the bounded total-variance enables projected data of the embedding manifolds to concentrate in arbitrarily narrow bands around the projected class mean, provided the ambient dimension is sufficiently large.

Gorban \& Tyukin’s stochastic separation theory characterizes intra-dataset probabilistic convexity: each sample is Fisher linearly separable from the rest of its dataset with high probability, provided the ambient dimension is sufficiently large. Our work extends this to inter-dataset separability between two distinct manifold distributions, proving high-probability linear separation with arbitrarily small classification error under bounded total-variance and projection non-singularity, provided the ambient dimension is large enough.

We develop a novel two-layer nested measure concentration analysis approach, applying Chebyshev’s inequality sequentially for dual-layer tail bounds and leveraging spherical projection vector concentration to integrate the concentration inequalities into one inequality. We should mention here that the stochastic separability theory not only reveals the stochastic separability properties in high dimensional spaces, but  provides a new mechanism for representation learning of large foundation models. To maximize cross-class manifold separability, one should minimize the intrinsic dimension of each class’s embedding manifold to reduce the  bounded total-variance.

There are a few perspective research directions related to the stochastic separability. First one is how to extend the stochastic separability theory to disentangled embedding manifolds for improved inference robustness. Second one is to investigate the optimal separating projection directions beyond random projections. Still other one is to estimate the minimal embedding dimension $n$, given $\epsilon$ and $\delta$.

\bibliographystyle{plain}  
\bibliography{reference}

\newpage
\begin{appendix}
\section{Fourth Moment Identities of Uniform Variables on $\mathbb S^{n-1}$ }

Let random vector $\mathbf{x} \in \mathbb S^{n-1}$ follow uniform distribution, satisfying:
\begin{equation}  \label{Normonex}
\sum_{i=1}^n x_i^2 =1.
\end{equation} 
Taking an expectation of both sides of the above equation, we have
\begin{equation}  \label{ENormonex}
\sum_{i=1}^n \mathbb{E}  [ x_i^2 ] =1.
\end{equation} 
Using the component symmetry $\mathbb{E} [x_i^2]=\mathbb{E} [x_j^2]$, we have
\begin{equation} 
\mathbb{E} [x_i^2]= \frac 1 n.
\end{equation}  
Define the fourth moment of $x_i$ as $ M_4=\mathbb{E}[x_1^4] $, and denote $V_{12}=\mathbb{E}[x_1^2 x_2^2] $. Squaring identity \eqref{Normonex}, and taking expectation on both sides of the equation, we obtain 
\begin{equation}  \label{MVrelation}
n M_4 + n(n-1) V_{12} =1, 
\end{equation}  
where we utilize the moment symmetry ($ M_4=\mathbb{E}[x_i^4], V_{12}=\mathbb{E}[x_i^2x_j^2], \ \forall i\neq j $). By rotational symmetry of the unit sphere, rotating coordinates $ (x_1,x_2) $ by $ \pi/4 $ preserves cross-moment $ V_{12} $:
\begin{equation} 
\mathbb{E} \left[ \left( \frac{x_1+x_2}{\sqrt 2}\right)^2 \left( \frac{x_1-x_2}{\sqrt 2}\right)^2 \right] = V_{12}.
\end{equation}  
Expand the left-hand side:
\begin{equation} 
\frac 1 4 \left( \mathbb{E} \left[ x_1^4 \right]+\mathbb{E} \left[ x_2^4\right] -2 \mathbb{E} \left[  x_1^2 x_2^2\right] \right) =V_{12}.
\end{equation} 
Solving the above equation leads to the following relation
\begin{equation}  
M_4= 3V_{12}.
\end{equation}  
Substituting the relation into \eqref{MVrelation}, we obtain the closed-form solutions for the moments:
\begin{equation}  
M_4=\frac{3}{n(n+2)} , \quad V_{12}= \frac{1}{n(n+2)} .
\end{equation} 
As ambient dimension $ n\to\infty $, both second and fourth component moments decay to zero.

\end{appendix}

\end{document}